\documentclass[letterpaper, 10 pt, conference]{ieeeconf}  
\IEEEoverridecommandlockouts                              
\usepackage{graphics}    
\usepackage{times}       
\usepackage{amsmath}     
\usepackage{amssymb}     
\usepackage{graphicx}
\usepackage{algorithm}
\usepackage[noend]{algpseudocode}
\usepackage{cite}
\usepackage{multirow}
\usepackage{subcaption}

\usepackage[font=small]{caption}

\def\eqref#1{Eq.~(\ref{#1})}

\usepackage{booktabs}
\usepackage{pifont}

\newcommand\etal{\emph{et al.}}

\def\argmax{\mathop{\rm argmax}}

\usepackage{comment}

\newif\ifshowappendix

\showappendixfalse

\ifshowappendix
  
\else
  \excludecomment{appendixsection}
\fi

 \title{\LARGE \bf Semantic- and Density-Aware Planning for \\ Accessibility-Preserving Multi-Object Placement} 

\author{Benno Wingender \and Nils Dengler \and Nicolas Busch \and Sicong Pan \and Maren Bennewitz
}

\begin{document}
\maketitle
\thispagestyle{empty} 
\pagestyle{empty}

\begin{abstract}

Long-term manipulation planning requires robots to reason not only about immediate task success but also about how current decisions affect future interactions with the environment.
In this context, household service robots may need to organize groceries in partially occupied shelves while using limited storage space efficiently and preserving access for subsequent placements.
In this paper, we consider an online multi-object shelf-placement setting in which future objects arrivals are unknown.
Existing approaches do not jointly address semantic organization, dense space utilization, and manipulator accessibility during sequential shelf filling.
To address this gap, we propose Semantic-Dense Placement Planning~(SDPP), an accessibility-preserving approach that ranks candidate poses using a semantic-density score combining inter-object semantic similarity with spatial proximity.
An Accessibility Map~(AM) further filters candidates unlikely to be reachable before motion planning and penalizes placements that reduce the remaining accessible workspace.
Simulation experiments show that SDPP significantly improves semantic placement quality over state-of-the-art baselines and achieves the highest average shelf density, while the AM substantially reduces the time required to identify feasible placement poses.
A qualitative real-world experiment demonstrates the applicability of our pipeline in a domestic shelf-storage scenario.

\end{abstract}

\section{Introduction}
\label{sec:intro}


Household service robots can assist users with everyday manipulation tasks, such as storing groceries. In this setting, a robot must handle objects sequentially and place them into a partially occupied, multi-level household shelf. At the time of each placement, however, the remaining items may still be occluded inside shopping bags or located outside the robot’s field of view. Each decision must therefore be made solely on the basis of the current shelf state and the object to be placed, without knowledge of the identities, arrival order, or geometries of future objects.
From the user’s perspective, the resulting arrangement should keep functionally or categorically related objects close to one another, use the available storage capacity efficiently, and preserve shelf regions that remain accessible for subsequent placements. The robot must therefore construct a dense and semantically organized shelf incrementally, while avoiding decisions that restrict future placement opportunities.

Achieving these objectives is challenging because the task is sequential: every placement changes both the remaining free space and the regions that can still be reached by the manipulator.
A pose that is collision-free for the current object may fragment the free workspace or block a narrow approach path, making  valid shelf regions unusable for subsequent placements.
The objectives can also conflict: semantic organization may favor a location near related objects, whereas dense space utilization and accessibility preservation may favor a different placement.
Unlike look-ahead bin-packing methods that work with simplified object geometries and top-down access~\cite{fang2026effective,zhao2022learning}, household objects arrive unpredictably and vary widely in shape.
The planner must therefore preserve accessible shelf space, since a placement that is locally suitable for the current object may reduce the feasibility and quality of the overall arrangement.

Existing work on robotic stowing and packing has advanced robust object manipulation and dense geometric storage~\cite{eppner2016lessons,morrison2018cartman, wang2021dense}, while shelf-replenishment and arrangement methods have addressed motion and grasp planning, retrieval efficiency, and arrangement stability~\cite{costanzo2021can, chen2022optimal, motoda2022shelf}. 
However, these approaches generally do not consider semantic shelf organization.
In contrast, semantic and preference-aware methods infer suitable placement targets from object relations, scene context, or user preferences~\cite{abdo2015collaborative, wu2023tidybot, wang2024apricot, ramachandruni2025personalized, adeleye2022putting}.
Although some of these methods account for geometric feasibility or environmental constraints, they do not jointly optimize semantic organization, dense space utilization, and manipulator accessibility as a confined shelf is sequentially filled.

\begin{figure}[t]
  \centering
 \includegraphics[width=\columnwidth, trim=0 30 0 0, clip]{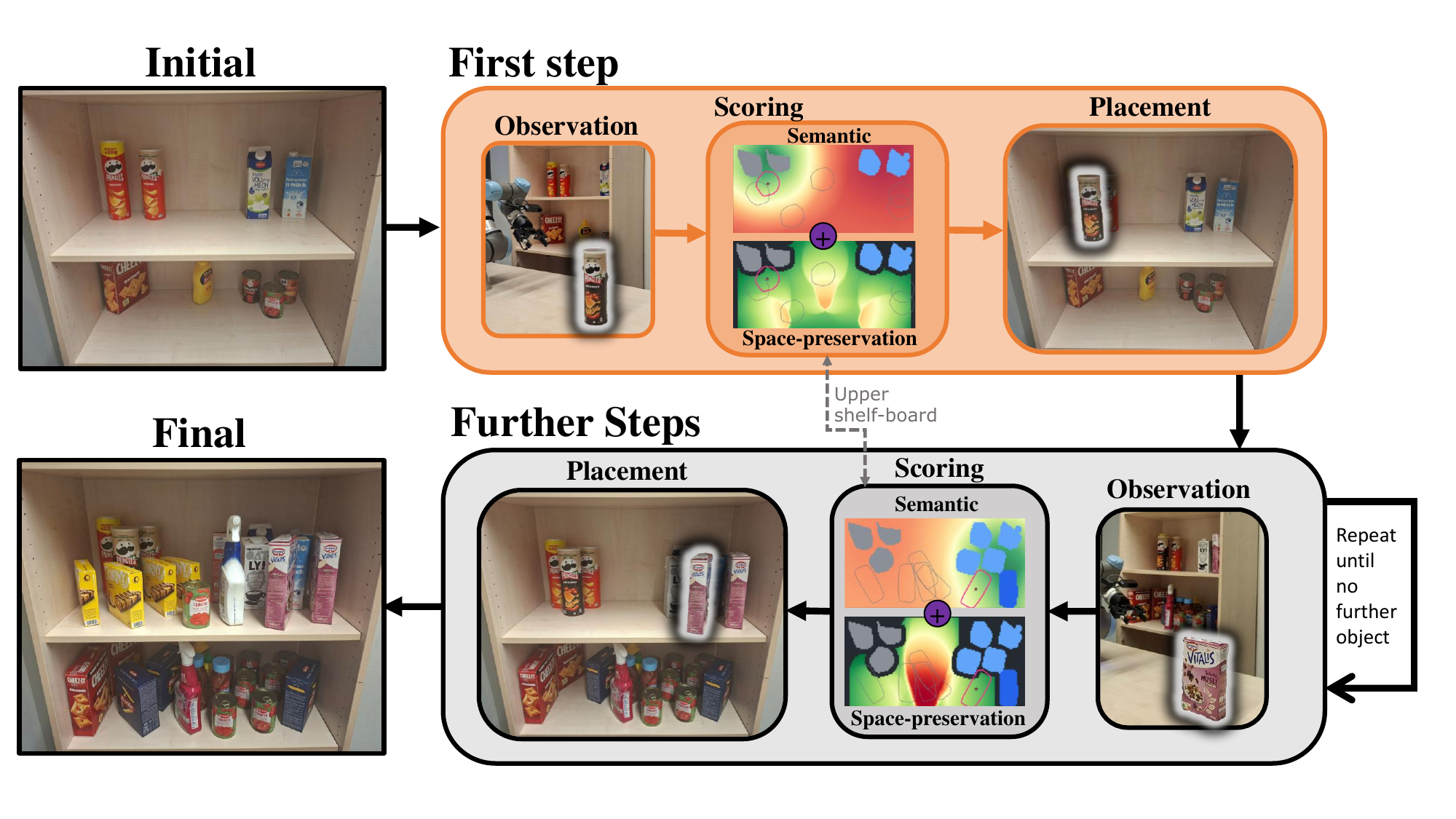}
\caption{Overview of the online multi-object shelf-placement task and the proposed approach.
Starting from a partially occupied multi-level shelf, objects are observed and placed one at a time without knowledge of future arrivals.
At each step, candidate placements on the selected shelf level are evaluated according to semantic-dense organization and accessibility preservation, after which the highest-ranked feasible placement is executed.
This process is repeated until no further object can be placed, producing a dense and semantically organized shelf arrangement.
}
  \label{fig:cover}
\vspace{-10px}
\end{figure}

To address this gap under the online setting considered here, we propose Semantic-Dense Placement Planning~(SDPP), an accessibility-preserving approach that jointly reasons about semantic organization and dense space utilization without relying on knowledge of future arrivals.
SDPP ranks candidate placement poses using a semantic-density score that combines inter-object semantic similarity with spatial proximity, encouraging compact local arrangements of related objects.
To preserve future placement opportunities, we introduce an Accessibility Map~(AM) that filters candidates unlikely to be reachable before motion planning and penalizes placements that substantially reduce the remaining accessible workspace.
Together, these components enable dense and semantically organized shelf arrangements while preserving accessibility as the shelf is progressively filled, achieving significantly higher semantic organization scores and greater shelf density than the baselines.
Figure~\ref{fig:cover} illustrates the considered task setting and the complete sequential placement process.

The main contributions of this paper are:
\begin{itemize}
    \item We propose a semantic-density placement score that combines semantic similarity with spatial proximity to form compact arrangements of related objects.
    \item We introduce an accessibility map that filters manipulation-inaccessible candidates and penalizes the loss of reachable workspace to  reduce the time required to identify feasible placement poses.
    \item We formulate an online shelf-placement setting with unknown future arrivals and integrate the proposed components into an accessibility-preserving SDPP pipeline.
\end{itemize}
Code of our pipeline will be made available upon publication.

\section{Related Work}
\label{sec:related}


\subsection{Object Stowing and Arrangement}

Robotic stowing has been studied extensively in logistics, particularly in the Amazon Robotics Challenge, where robots manipulated diverse objects in storage pods and bins~\cite{eppner2016lessons,morrison2018cartman}.
Related robotic bin-packing methods aim to maximize geometric storage efficiency in bounded containers, often relying on vertical stacking and gravity-supported arrangements~\cite{pantoja2024comprehensive,dahmani2025reinforcement,ali2022line,wang2021dense}.
Recent industrial systems similarly focus on storage density and throughput~\cite{hudson2025stow}.

Shelf storage differs from these settings because objects must be placed on accessible support surfaces in confined spaces.
Work on supermarket shelf replenishment has therefore primarily addressed robust manipulation strategies~\mbox{\cite{winkler2016knowledge,koutras2025robotic}}, while other shelf-arrangement methods optimize objectives such as retrieval cost~\cite{chen2022optimal}, arrangement stability, and insertion feasibility~\cite{motoda2022shelf,chen2023predicting}.
These approaches address important geometric and manipulation challenges, but generally do not jointly consider semantic organization, dense space utilization, and spacial accessibility.

\subsection{Semantic Placements}

Semantic placement methods infer suitable placement locations from object relations, scene context, or user preferences~\cite{abdo2015collaborative, wu2023tidybot,wang2024apricot}.
Ramachandruni~\etal~\cite{ramachandruni2025personalized}, for example, proposed ContextSortLM, which infers personalized organization preferences from previous observations and the current partial arrangement.
Other approaches use foundation models to reason about contextually appropriate placements from visual observations or natural-language prompts~\cite{zhao2025anyplace,kapelyukh2024dream2real,lee2024spots}, while classical word embeddings provide fixed representations for quantifying semantic relationships between object labels~\cite{pennington2014glove}.
Most closely related to our work, ~\mbox{Adeleye~\etal~\cite{adeleye2022putting}} combine hierarchical grocery categories with language embeddings and greedily place each incoming item near the most semantically related object.
However, existing semantic placement methods primarily address target or reference-object selection for individual placements, rather than how repeated placement decisions affect dense space utilization and manipulator accessibility as a confined shelf is progressively filled.

\begin{figure*}[t]
\vspace{-2px}
    \centering
    \includegraphics[width=\linewidth, trim=0 170 0 180, clip]{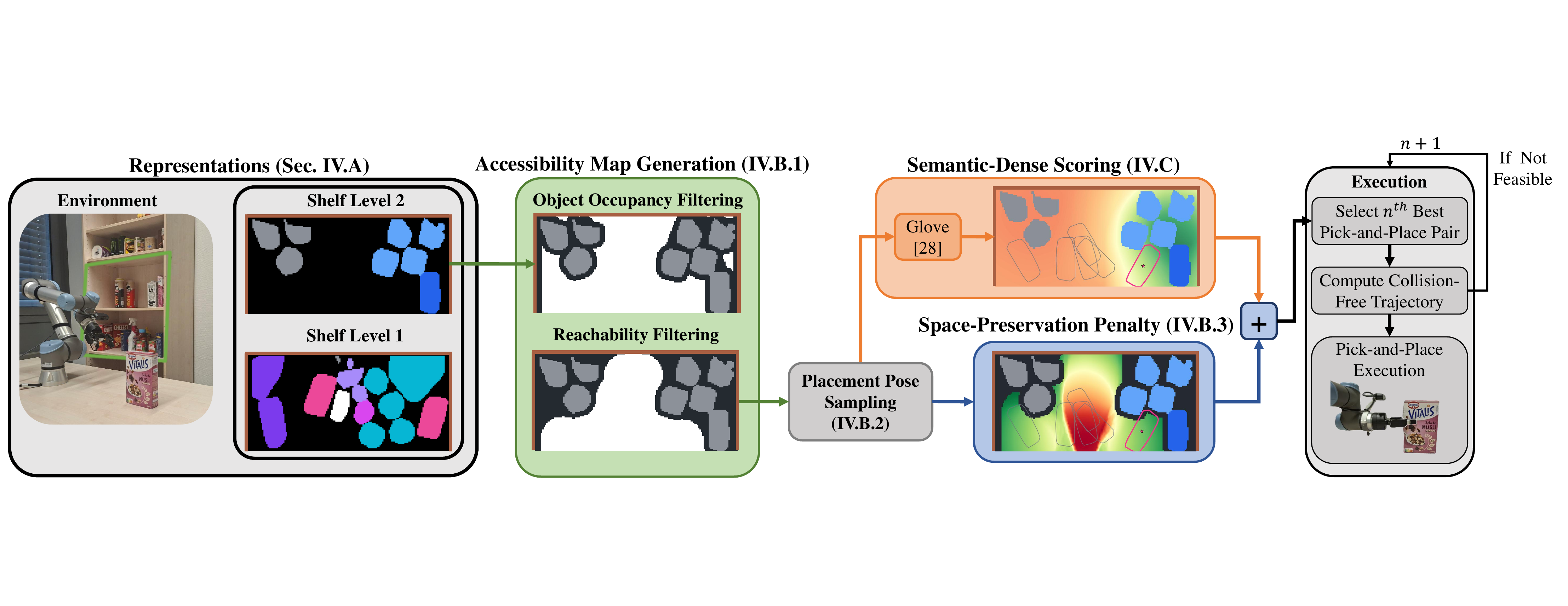}
    \caption{Overview of our semantic-dense placement planning pipeline. The perceived shelf environment is converted into 2D~occupancy map for each shelf level (left), from which the Accessibility Map (AM) identifies occupied (gray) and manipulation-inaccessible (black) regions~(green panel). 
    Candidate placement poses are sampled in the remaining workspace and evaluated using a semantic-density score, which combines \mbox{GloVe-based~\cite{pennington2014glove}} object similarity with spatial proximity (orange panel), and a space-preservation penalty~(blue panel) that estimates the loss of accessible shelf space caused by each placement pose. 
    The combined score ranks candidate placements poses. pairs release and grasp pose candidates, and then sequentially checks for a collision-free pick-and-place trajectory (right).
    }
    \label{fig:Overview}
\vspace{-8px}
\end{figure*}

\section{Problem Definition}
Storing multiple household items on multi-level domestic shelves requires a robot to place objects in partially filled shelves while balancing two competing objectives: (i)~semantic arrangement and (ii)~dense space utilization.
Here, semantic arrangement favors objects of similar classes to be stored spatially close, while dense space utilization aims to maximize the shelf-space utilization regardless of the object type.
Due to unknown object properties and incomplete meshes, stacked placements may be unstable, so we consider only collision-free placements on shelf boards.
Formally, let the shelf state $\mathcal{S}_t$ at time step $t$ contain a set of already placed objects $\mathcal{O}_t=\{o_{1},\cdots,o_{n}\}\subseteq \hat{\mathcal{O}}$ with poses~\mbox{$x_{i}\in SE(3)$}, where $\hat{\mathcal{O}}$ denotes the unknown set of all objects.
Furthermore, let \mbox{$\hat{o} \in \hat{\mathcal{O}}\setminus \mathcal{O}_t$} be the upcoming object that needs to be placed by the robot inside the shelf.
The set of semantic classes $\mathcal{C}$ over all objects $\hat{\mathcal{O}}$ is a priori known, while the exact geometry~$g(\hat{o})$ of objects becomes observable only upon arrival.
For two objects~$o_i$ and~$o_j$ already placed within the shelf, we denote \mbox{$s(o_i,o_j)\in[-1,1]$} as the semantic similarity, where larger values indicate higher semantic relation.
The robot must select a collision-free and robot-reachable placement pose $x_{\hat{o}}^{*}\in SE(3)$ that maximizes
\begin{equation}
    x_{\hat{o}}^{*}
=
\argmax_{x \in SE(3)}
\left[
\lambda_1 J_{\mathrm{sem}}(x \mid \hat{o})
+
\lambda_2J_{\mathrm{space}}(x \mid\hat{o})
\right],
\end{equation}
where $J_{\mathrm{sem}}$ quantifies the semantic organization objective based on the semantic similarities $s(\hat{o},o_i)$, $J_{\mathrm{space}}$ quantifies the shelf-space utilization objective, and $\lambda_{1,2}\in[0,1]$ controls the trade-off between these two objectives.

\section{Our Approach}
\label{sec:main}
We formulate placement planning as the optimization of a unified score that balances semantic consistency, spatial compactness, and long-term accessibility.
As illustrated in Fig.~\ref{fig:Overview}, we first introduce our environmental representation, followed by the Accessibility Map~(AM), which estimates reachable shelf regions, rejects placement candidates that are unlikely to be executable, and penalizes placements that substantially reduce the remaining accessible workspace through a space-preservation penalty.
We then introduce a semantic-density score that combines semantic similarity with spatial proximity to form compact arrangements of related objects while separating dissimilar ones.
Finally, these components are integrated into a sequential manipulation pipeline that executes the highest-scoring feasible placement.

\subsection{Environmental Representation}\label{ssec:env}
We model each shelf board as a 2D occupancy grid $\mathcal{S}^l_t$ by projecting each object onto the shelf plane as a 2D footprint and additionally store the object's height, center position, and semantic class, as illustrated in the left panel of Fig.~\ref{fig:Overview}.

\subsection{Accessibility Map}

Dense shelf placement requires using free space efficiently without blocking regions that may be needed for subsequent objects. 
However, geometric collision checks alone do not indicate whether a placement pose is reachable by the manipulator or whether placing an object there would obstruct access to larger free regions.
To address this, we introduce the Accessibility Map~(AM) to serve two purposes. 
First, it guides placement sampling towards regions that are likely reachable by the manipulator, avoiding expensive motion planning for candidates that are unlikely to be executable. 
Second, it estimates how much reachable workspace would be lost after a hypothetical placement, allowing the planner to penalize placements that unnecessarily reduce future accessibility while still permitting dense object arrangements.

\subsubsection{\textbf{Map Generation}}\label{AM_mapgen}
To efficiently approximate which shelf regions remain accessible for placement, we construct the AM from the footprints of the objects already present on the shelf. 
Moreover, we represent the accessible workspace as a binary mask, initially marking only the projected object footprints as inaccessible. 
Since placements directly adjacent to or behind existing objects are difficult to execute without risking collisions, we dilate each object footprint and the shelf walls and add a rear-facing cone for each object.
The resulting mask, shown in black in the upper panel of the Accessibility Map Generation block in Fig.~\ref{fig:Overview}, represents regions in which candidate object centers should not be placed.
Moreover to account for the clearance required by the gripper during placement, we apply morphological closing to the dilated occupancy mask using an elliptical structuring element derived from the gripper width, illustrated in the lower panel of the AM Generation block in Fig.~\ref{fig:Overview}.

\subsubsection{\textbf{Object Placement Sampling}}\label{ssec:sampling}
For each new object~$\hat{o}$, we sample a set of candidate placement poses on each shelf board based on the object's 2D footprint and 12 possible orientations.
In dense shelf environments, many geometrically collision-free candidates remain inaccessible to the manipulator.
We therefore use the AM to reject unlikely candidates before motion planning.
Specifically, we introduce two AM-based rejection strategies, denoted \emph{Footprint-Constrained} (FC) and \emph{Center-Constrained}~(CC). 

In footprint-constrained sampling, a candidate placement is rejected if any cell of its projected footprint overlaps the inaccessible region. This conservative strategy restricts sampling to poses whose entire footprint lies within the estimated accessible area. 
In contrast, In center-constrained sampling, a candidate is rejected only if its footprint center lies inside the inaccessible region or if the footprint overlaps cells occupied by existing objects or the shelf boundary. 
This less restrictive strategy permits placements near the boundary of inaccessible regions and supports denser arrangements while maintaining a high probability of successful execution.
We first apply FC and fall back to the less restrictive CC strategy if no executable placement is found.
Candidate poses are sampled independently on each shelf board until either~$n$~$(n=250)$ valid placements have been found or no additional feasible candidates can be generated.

\subsubsection{\textbf{Space-Preservation Penalty}}\label{AM_spatialblock}
While the AM rejects placement poses that are unlikely to be executable, it is also important to distinguish between feasible placements that preserve future placement opportunities and those that unnecessarily block large portions of the remaining workspace.
Therefore, for every candidate placement we recompute the AM under the hypothetical shelf configuration and measure the loss of reachable area.
This is done by calculating the amount of cells being inaccessible before $|U(\mathcal{S}_t)|$ and after the hypothetical placement $|U(\mathcal{S}_{t}^i)|$ with:
\begin{equation}
    \mathcal{V}_{\mathrm{AM}} = \frac{\mid U(\mathcal{S}_{t}^i)\mid - \mid U(\mathcal{S}_t)\mid}{\mid \mathcal{S} \mid},
\end{equation}
where $\mid \mathcal{S} \mid$ is the size of the shelf.
Thus, candidate placements that substantially reduce the remaining accessible workspace receive a higher penalty.

\subsection{Local Semantic-Density Scoring}
\label{semantic_score}
\begin{figure}
    \centering
    \includegraphics[
        width=0.8\linewidth,
        trim={0mm 5mm 0mm 17mm},
        clip
    ]{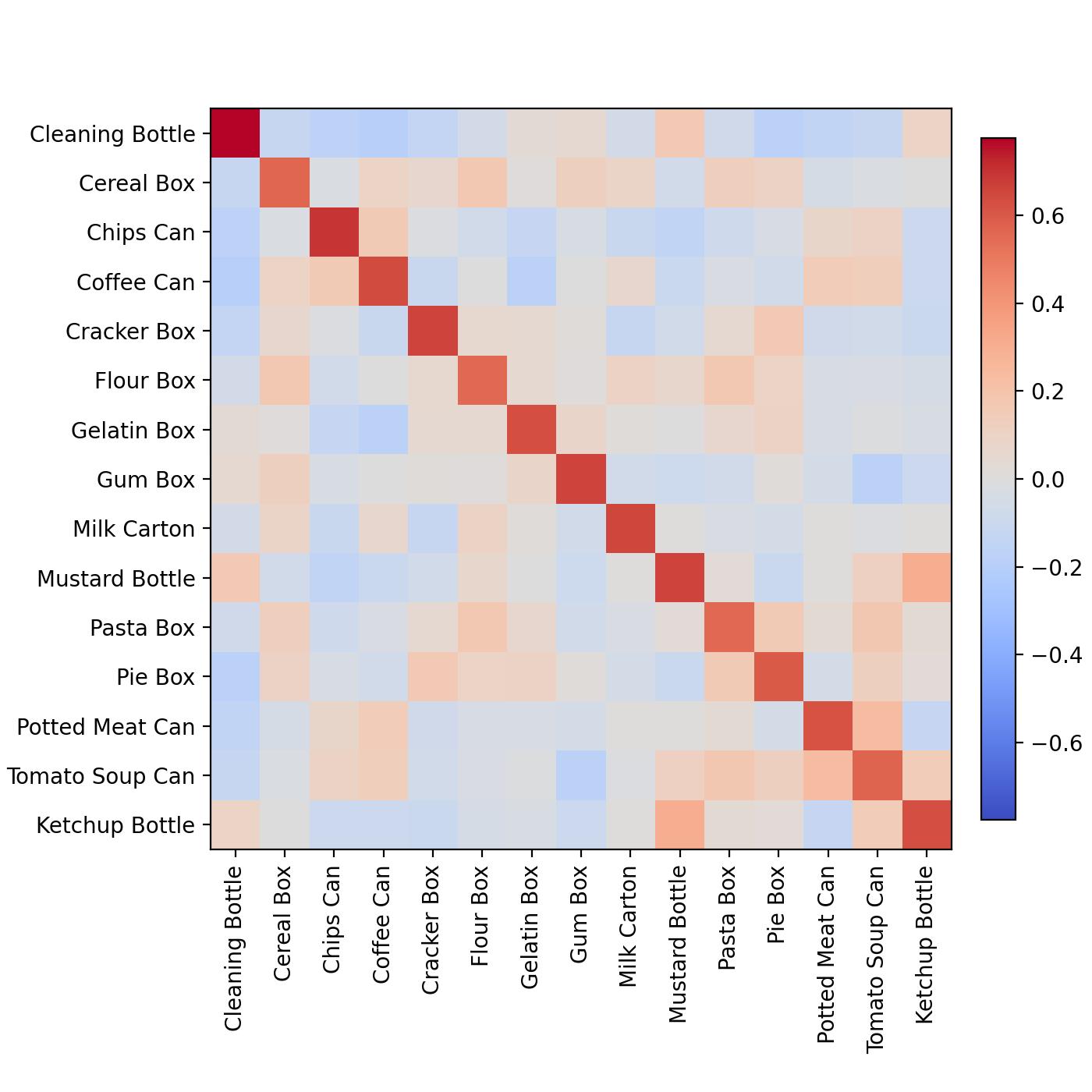}
    \vspace{-8px}
    \caption{Pairwise semantic similarity matrix computed from GloVe 
    embeddings~\cite{pennington2014glove} after contrast normalization. Warmer colors indicate semantic similarity, while cooler colors denote dissimilarity. 
    }
    \label{fig:similarity}
    \vspace{-14px}
\end{figure}

To generate semantically meaningful shelf arrangements, objects should be placed close to semantically related neighbors. At the same time, semantic relationships should primarily influence the local neighborhood rather than the entire shelf. We therefore combine semantic similarity with spatial proximity, encouraging compact local object clusters without enforcing a global shelf organization.

To quantify semantic relationships, we compute pairwise similarity scores
$s(o_i,o_j)\in[-1,1]$ using GloVe~word embeddings~\cite{pennington2014glove}. Since object labels typically contain both semantic information (e.g., \textit{mustard}) and object form (e.g., \textit{bottle}), we split each label into semantic words \mbox{$(\mathbf{v}(w_{i,1}),\ldots,\mathbf{v}(w_{i,n_i}))$} with combined embedding \mbox{$\mathbf{m}_i=\sum_{w\in M_i}\mathbf{v}(w)$}. Additionally we also use the full form word~$f_i$ with its embedding \mbox{$\mathbf{f}_i=\mathbf{v}(f_i)$}.
The similarity between two objects is then defined as
\begin{equation}
\begin{aligned}
s(o_i,o_j)
&=
\alpha\cos(\mathbf{m}_i,\mathbf{m}_j)
+
(1-\alpha)\cos(\mathbf{f}_i,\mathbf{f}_j),\\
&\text{where}\quad
\cos(\mathbf{a},\mathbf{b})
=
\frac{\mathbf{a}^{\top}\mathbf{b}}
{\|\mathbf{a}\|_2\|\mathbf{b}\|_2}.
\end{aligned}
\end{equation}
Throughout this work, we use $\alpha=0.8$, giving higher importance to semantic content while still accounting for object form.
Since pantry and household objects are often assumed to be semantically related, the raw similarity values exhibit only limited contrast.
We therefore center the similarity matrix by subtracting, for each object pair, the average similarity of both objects to all other object classes:
\begin{equation}
\resizebox{\dimexpr\linewidth-3.5em\relax}{!}{$\displaystyle
\hat{s}(o_i,o_j)
=
s(o_i,o_j)
-
\frac{1}{2(|\mathcal{C}|-1)}
\left(
\sum_{k\in\mathcal{C}\setminus \{i\}} s(o_i,o_k)
+
\sum_{l\in\mathcal{C}\setminus \{j\}} s(o_j,o_l)
\right)
$}
\label{eq:normalized_similarity}
\end{equation}
The resulting similarity matrix is shown in Fig.~\ref{fig:similarity}.
Rather than considering semantic similarity alone, we additionally weight each neighboring object according to its spatial distance from the candidate placement. Consequently, nearby objects have the strongest influence on the placement decision, encouraging compact local groupings, while distant objects contribute only weakly to preserve the broader shelf context. For a candidate placement position $\mathbf{x}$ of object $\hat{o}$, we define
\begin{equation}\label{math:semantic}
\resizebox{\dimexpr\linewidth-3.5em\relax}{!}{$\displaystyle
\mathcal{V}_{\mathrm{SemDen}}(\mathbf{x}\mid \hat{o})
=
\sum_{o_j\in\mathcal{O}_t^l}
\hat{s}(\hat{o},o_j)
\max\!\left(
0,
1-\frac{d(\mathbf{x},o_j)}{d_{\max}}
\right),
$}
\vspace{-3px}
\end{equation}
where $\mathcal{O}_t^l$ denotes the objects already placed on shelf level~$l$, $d(\mathbf{x},o_j)$ is the shortest 2D Euclidean distance between the candidate position and the footprint of object $o_j$, and $d_{\max}$ specifies the maximum distance over which neighboring objects influence the score. 

\subsection{Placement Scoring}
The accessibility map and semantic score capture complementary objectives. While the semantic score encourages compact arrangements of semantically related objects, the Accessibility Map preserves future placement opportunities by penalizing placements that unnecessarily reduce the remaining reachable workspace. We combine both objectives into the final placement score
\begin{equation}\label{math:semantic_score}
\mathcal{V}_{\mathrm{p}}(\mathbf{x}\mid \hat{o})
=
\mathcal{V}_{\mathrm{SemDen}}(\mathbf{x}\mid \hat{o})
-
w_2\mathcal{V}_{\mathrm{AM}}(\mathbf{x}\mid \hat{o}),
\end{equation}
where $w_2$ controls the trade-off between semantic organization and accessibility preservation.

\subsection{Pick-and-Place Execution}\label{ssec:pipeline}
To execute these placements on a real robot, we couple placement planning with grasp generation and motion planning. 
For each incoming object, we first generate AM-feasible candidate placements, rank them according to $\mathcal{V}_{\mathrm{p}}$, and subsequently validate them using grasp and motion planning until an executable placement is found.
To generate grasp candidates, we build upon GoalGrasp~\cite{gui2025goalgrasp} and generate grasp candidates from an object bounding box. However, instead of fitting the bounding box directly in 3D, we first project the object point cloud onto the supporting plane, compute the minimum-area bounding rectangle of its convex hull, and subsequently lift it back into 3D.
Furthermore, inspired by Wingender~\etal~\cite{wingender2025generalized}, 
all grasp poses are represented in the object's local coordinate frame, allowing them to be directly transformed to every candidate placement pose.
Finally, we execute the selected grasp-place pair using NVIDIA's cuRobo motion planner~\cite{sundaralingam2023curobo}. 

\section{Experimental Evaluation}
\label{sec:exp}

We evaluate our SDPP in simulation and on a real robotic platform to answer three questions:
(i) Does SDPP produce more semantically organized and space-efficient shelf arrangements, 
and how do the individual scoring terms contribute to this performance?
(ii) Can AM-based filtering reduce the IK and motion-planning time required to identify an executable placement without excessively pruning feasible candidates?
(iii) Can the complete SDPP pipeline be applied to a realistic shelf-storage task on real hardware?

\subsection{Experimental Setup}

\subsubsection{Platforms and Objects}
Simulation experiments were conducted in PyBullet using a Neobotix MMO-700 mobile manipulator.
Small adjustments of the mobile base enabled the robot to access all three shelf levels, allowing evaluation on fully populated shelves.
Real-world experiments were performed using a stationary UR5e 6-DoF manipulator equipped with a Robotiq 2F-85 gripper and a wrist-mounted Orbbec Gemini 336 RGB-D camera.
Experiments were run on a workstation equipped with an NVIDIA RTX~3080~Ti GPU, an Intel Core i9-11900K CPU, and 64\,GB RAM.
The evaluation in Sec.~\ref{ssec:URM_sampling} was conducted on a workstation with an NVIDIA RTX~4060 GPU, an Intel Core i7-14700HX CPU, and 32\,GB RAM.
To aid robust execution, we filter grasp candidates based on gripper width, clearance, height, and rotation, and subsequently cluster them.
To avoid overestimating inaccessible space, we reduce the inaccessible region by $10\,\mathrm{cm}$ along the shelf depth before sampling.
As we focus on placement planning, we assume successful execution of the selected force-closure grasp.

\subsubsection{Initial Shelf Configuration Generation}
To evaluate the placement methods under diverse and plausible initial conditions, we generate partially occupied shelves using a probabilistic sampling method~\cite{marques25rss} that promotes local semantic groupings while respecting object footprints, shelf boundaries, and occupancy constraints.
We extend the method to multi-level shelves by allowing objects of the same semantic class to be distributed across different shelf levels with a $5\%$ probability.
Sampling continues until the specified initial object count or shelf occupancy is reached.
All evaluated methods use the same initial configurations and incoming-object sequences to enable paired comparisons.

\begin{figure*}[t]
    \centering

    \includegraphics[width=0.55\textwidth,
    trim={0mm 3mm 0mm 3mm},
        clip]
        {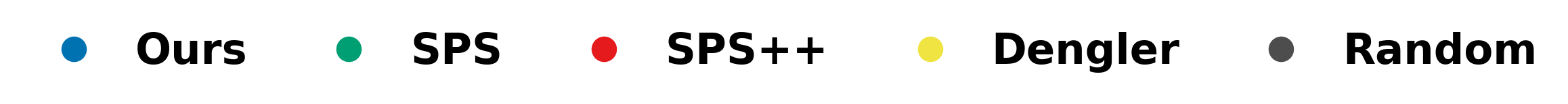}
    \par\vspace{2mm}

    \vspace{-8px}
    \begin{subfigure}[t]{0.32\textwidth}
        \centering
        \includegraphics[width=\linewidth]
            {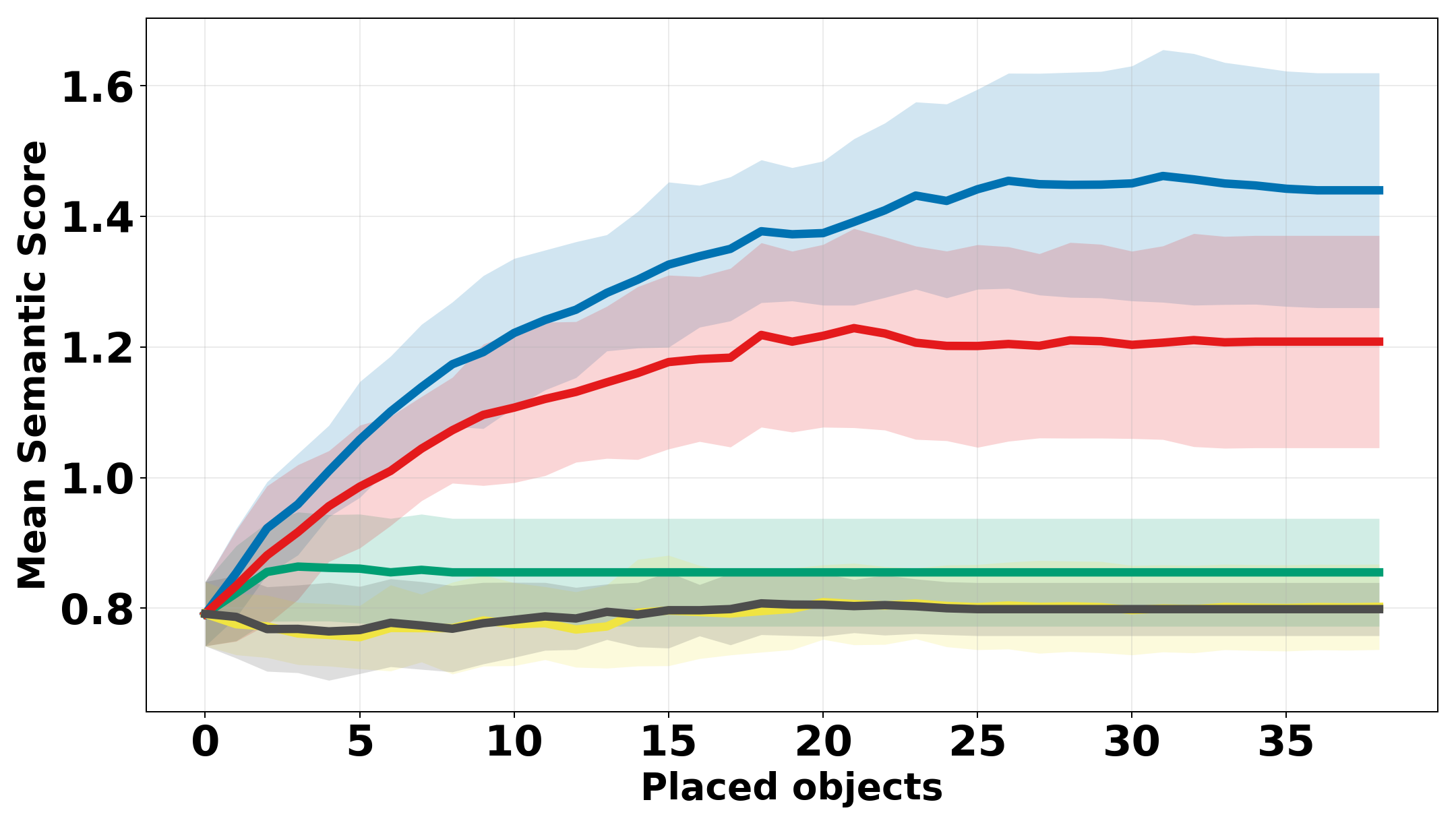}
        \vspace{-16px}
        \caption{Semantic Score: Mean semantic similarity score over all objects within $25\,\mathrm{cm}$.}
        \label{fig:semantic_eval_25cm}
    \end{subfigure}
    \hfill
    \begin{subfigure}[t]{0.32\textwidth}
        \centering
        \includegraphics[width=\linewidth]
            {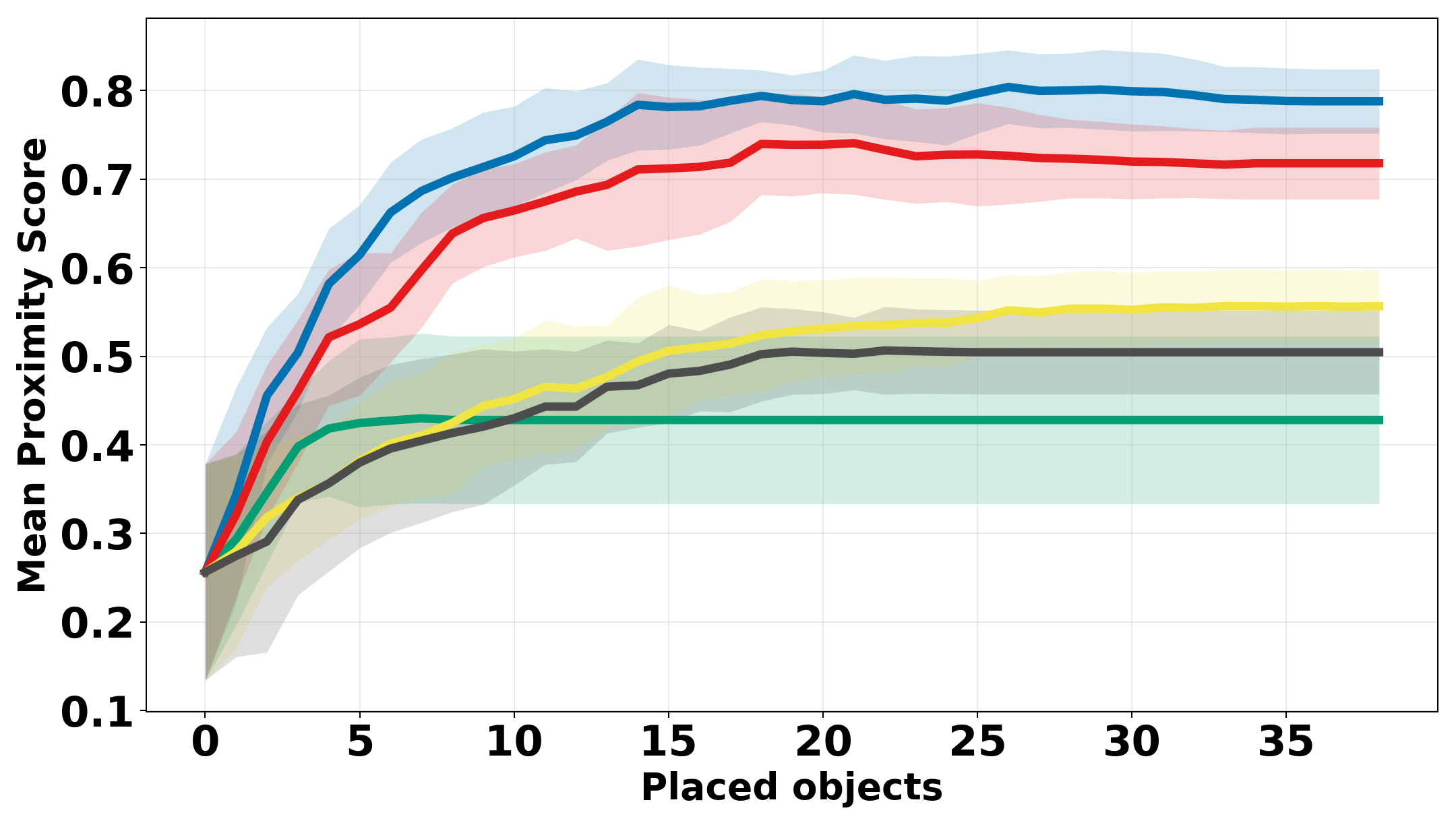}
        \vspace{-16px}
        \caption{Proximity Score: Mean semantic similarity score to the most related object within $10\,\mathrm{cm}$.}
        \label{fig:semantic_eval_10cm}
    \end{subfigure}
    \hfill
    \begin{subfigure}[t]{0.32\textwidth}
        \centering
        \includegraphics[width=\linewidth]
            {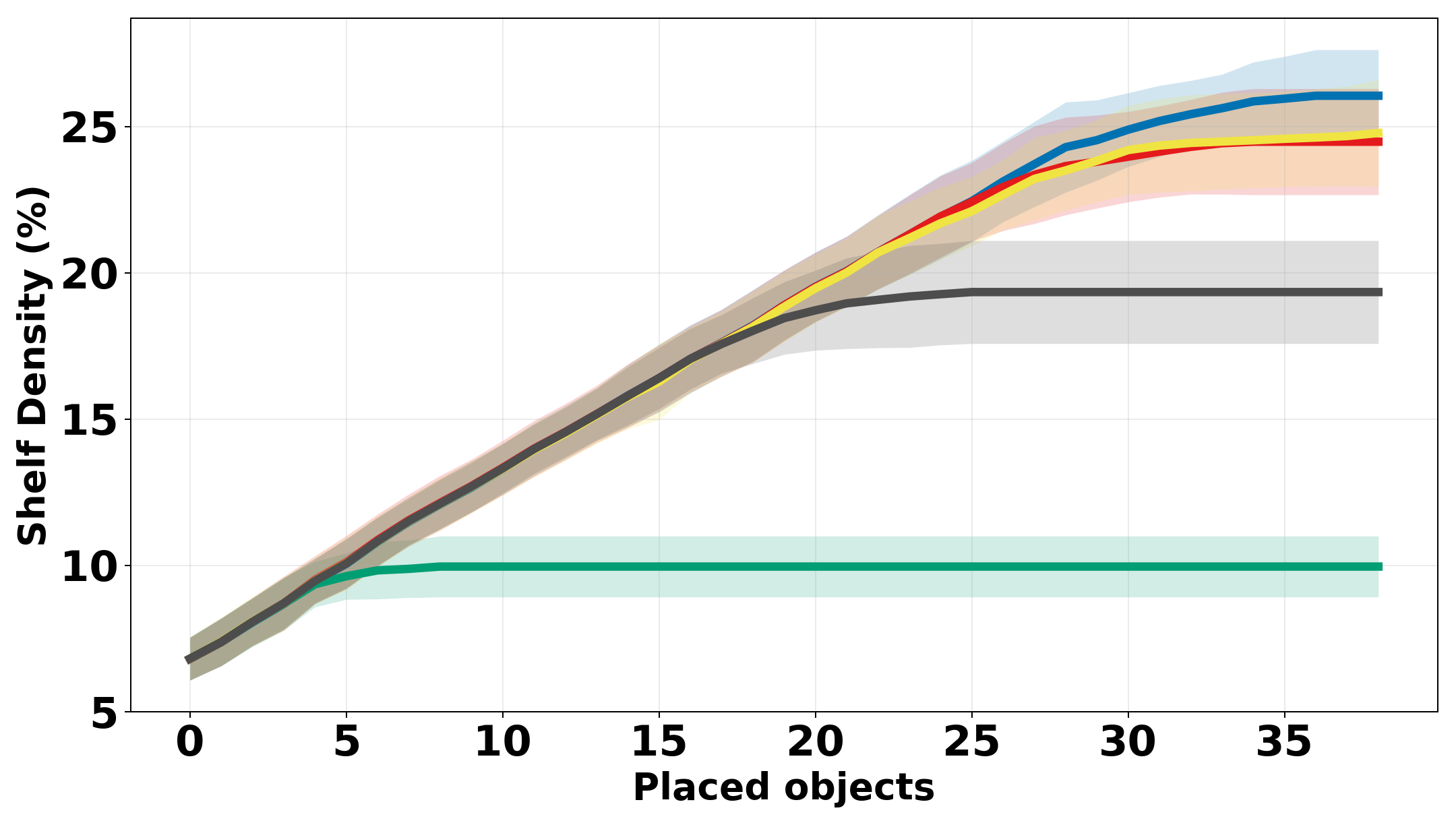}
        \vspace{-16px}
        \caption{Density Score: Mean occupied space divided by total space.}
        \label{fig:semantic_eval_density}
    \end{subfigure}
\vspace{-2px}
    \caption{
   Evolution of shelf-arrangement metrics over successive placements, averaged over 10 paired trials.
   Each trial starts with 12 objects, denoted as placement step~0, and terminates when no executable placement can be found for the current object.
   Panels~(a)--(c) report the semantic score, proximity score, and shelf density, respectively.
   SDPP maintains the strongest semantic organization as the shelf is filled while achieving shelf utilization comparable to the strongest density-oriented baselines.
    }
    \label{fig:semantic_eval}
\end{figure*}
\begin{table*}[t]
\centering
\small
\vspace{-8px}
\setlength{\tabcolsep}{4pt}
\resizebox{0.8\textwidth}{!}{%
\begin{tabular}{lccccc}
\toprule
\textbf{Variant}
& \textbf{Add. Obj. Placed} $\uparrow$
& \textbf{Semantic Sum} $\uparrow$
& \textbf{Semantic} $\uparrow$
& \textbf{Proximity} $\uparrow$
& \textbf{Density} $\uparrow$
\\
\midrule

Ours
& $\mathbf{30.7 \pm 3.9}$
& $\mathbf{61.6 \pm 11.5}$
& $\mathbf{1.440 \pm 0.180}$
& $\mathbf{0.788 \pm 0.036}$
& $\mathbf{0.261 \pm 0.016}$
\\

SPS~\cite{adeleye2022putting}
& $4.7 \pm 1.5$ 
& $14.1 \pm 1.9$ 
& $0.855 \pm 0.083$ 
& $0.428 \pm 0.095$ 
& $0.100 \pm 0.010$ 
\\

SPS++~\cite{adeleye2022putting}
& $28.2 \pm 3.6$ 
& $48.7 \pm 10.0$ 
& $1.208 \pm 0.163$ 
& $0.718 \pm 0.041$ 
& $0.245 \pm 0.018$ 
\\

Dengler~\cite{dengler2026learning}
& $28.6 \pm 4.6$ 
& $32.4 \pm 4.4$ 
& $0.802 \pm 0.065$ 
& $0.557 \pm 0.042$ 
& $0.248 \pm 0.018 ^{\dagger}$
\\

Random
& $19.9 \pm 2.7$ 
& $25.4 \pm 3.1$ 
& $0.799 \pm 0.041$ 
& $0.505 \pm 0.047$ 
& $0.193 \pm 0.018$ 
\\

\midrule
\multicolumn{6}{l}{\textbf{Ablations}} \\

$\mathrm{Ours}_{(w_{2}=0)}$
& $29.2 \pm 2.7$
& $58.7 \pm 7.3$
& $1.429 \pm 0.148$
& $0.806 \pm 0.041$
& $0.250 \pm 0.010$
\\

$\mathrm{Ours}_{(w_{2}=5)}$
& $31.1 \pm 3.5$
& $55.7 \pm 10.1$
& $1.295 \pm 0.183$
& $0.727 \pm 0.076$
& $0.262 \pm 0.023$
\\

\bottomrule
\end{tabular}%
}

\vspace{-4px}
\caption{
Comparison of SDPP with the evaluated baselines and ablations.
Each shelf initially contains 12 objects; \emph{Add. Obj. Placed} denotes the number of additional objects successfully placed.
All methods are evaluated over 10 paired trials using the same initial shelf configurations and incoming-object sequences.
Values are reported as mean $\pm$ standard deviation, and bold indicates the best result among the main methods.
The semantic sum is computed over all objects in the final shelf, whereas the semantic, proximity, and density scores are averaged per object.
Paired Wilcoxon signed-rank tests show significant differences between SDPP and each baseline for all reported metrics ($p<0.05$), except for shelf density compared with Dengler~\cite{dengler2026learning}, marked by $^{\dagger}$.
No statistical significance tests are reported for the ablation variants.
Among the main methods, SDPP achieves the strongest overall performance, combining the best semantic organization with the highest placement capacity and average shelf density.
The ablations show that increasing the accessibility-preservation weight improves placement capacity and density, but eventually trades off semantic quality.
}
\label{tab:placement_metrics}
\vspace{-8px}
\end{table*}

\subsubsection{Shelf-Arrangement Evaluation Protocol}
We evaluate whether SDPP produces semantically organized and space-efficient shelf arrangements while preserving capacity for subsequent placements.
Each simulated trial starts from one of the generated initial configurations, containing 12 objects, with four objects on each of the three shelf levels.
At each placement step, the next object is randomly selected from the evaluation set and presented on a nearby side table for perception.
The planner observes only the current object; the remaining arrival sequence is not revealed.
A trial terminates when no executable placement can be found for the current object.
We conduct 10 paired trials for each method.
All methods are evaluated using the same initial shelf configurations and incoming-object sequences, ensuring that differences are attributable to the placement strategy rather than variations in the evaluated scenarios.

\subsubsection{Baselines}
We compare SDPP against four baselines spanning diverse placement strategies.
\paragraph*{\textbf{Semantic Proximity Sampling (SPS)}}
We reimplement the semantic grocery-placement method of Adeleye~\etal~\cite{adeleye2022putting} and denote it as \emph{Semantic Proximity Sampling~(SPS)}.
SPS considers reference objects in descending order of semantic similarity and places the incoming object at a distance determined by half the reference-object width, half the candidate-object width, and a $1\,\mathrm{cm}$ clearance.
To improve placement success, we evaluate a $3\times3$ grid of lateral and depth offsets spaced by $2.5\,\mathrm{cm}$, with 12 orientations.
\paragraph*{\textbf{Semantic Proximity Sampling++ (SPS++)}}
The original SPS strategy considers placements only to the left and right of the most semantically related object.
We therefore introduce a more spatially flexible variant, denoted as~\emph{SPS++}.
It uses the same candidate placements as SDPP but ranks each pose according to the highest semantic similarity to any object within a radius $d_{\mathrm{rad}}$.
We set \mbox{$d_{\mathrm{rad}}=10\,\mathrm{cm}$}, as this value achieved the best overall performance among the tested values of $5, 10$ and  $15\,\mathrm{cm}$.
SPS++ thereby isolates the effect of the proposed semantic-density score.
\paragraph*{\textbf{Density-driven placement}}
As a density-driven baseline, we use a clearance-based scoring~\cite{dengler2026learning}, i.e.,
for each candidate, we compute the minimum clearance to the objects already stored in the shelf.
Candidates with negative clearance are rejected, while collision-free clearances up to \mbox{$\tau=3\,\mathrm{cm}$} receive the maximum score.
For larger clearances, the score decreases exponentially up to a maximum considered distance of $25\,\mathrm{cm}$.
This strategy favors compact placements near existing objects but does not explicitly consider semantic organization or the preservation of accessible workspace.
\paragraph*{\textbf{Random placement}}
Finally, we include a random baseline that assigns equal scores to all candidate poses, resulting in a randomized evaluation order.
To isolate the effect of pose ranking, this baseline uses the same candidate-generation filter as SDPP.

\subsubsection{Metrics}
We evaluate the generated arrangements using three arrangement-quality metrics.
In addition, we report the number of successfully placed objects and an aggregate semantic sum.
The metrics are computed for the initial configuration and after every successful placement, allowing us to analyze both their evolution over time and final values.
\paragraph*{\textbf{Additional objects placed}}
We report the number of objects successfully placed beyond the 12 objects initially present in the shelf.
This directly measures how effectively a method preserves capacity for subsequent placements.
\paragraph*{\textbf{Semantic sum}}
We sum the per-object semantic scores over all objects in the final shelf configuration.
Unlike the semantic score described below, this aggregate measure reflects both the semantic quality of the arrangement and the number of objects successfully placed.
\paragraph*{\textbf{Semantic score}}
For each object in the shelf, we compute the proximity-weighted semantic value defined in Eq.~\ref{math:semantic} and report the mean over all objects.
High values indicate that objects are placed close to semantically related neighbors, whereas nearby dissimilar objects and isolated objects reduce the score.
\paragraph*{\textbf{Proximity score}}
For direct comparison with Adeleye~\etal~\cite{adeleye2022putting}, we determine for each placed object the highest semantic similarity to another object within $d_{\mathrm{rad}}=10\,\mathrm{cm}$ and report the mean over all objects.
This metric measures whether each object has at least one semantically related neighbor in its local vicinity.
\paragraph*{\textbf{Density score}}
At each placement step, we measure shelf-space utilization as the fraction of the total shelf-board area occupied by the projected object footprints across all shelf levels.
Higher values indicate denser and more space-efficient use of the available storage space.

\subsection{Shelf-Arrangement Performance}
\label{ssec:arrangement_results}
We first examine the overall performance of SDPP in producing good placement sequences.

\subsubsection{Semantic Organization}
As shown in Fig.~\ref{fig:semantic_eval}, SDPP consistently achieves the highest semantic scores among the evaluated methods.
The difference becomes pronounced after approximately five additional placements and continues to increase as the shelf is progressively filled.
These results indicate that SDPP increasingly improves local semantic organization as more objects are placed, despite the randomized arrival order of semantic classes.
SPS~\cite{adeleye2022putting} achieves semantic scores comparable to those of the more spatially flexible SPS++ variant during the early placement stages, but exhausts its feasible placement candidates substantially earlier.
This suggests that selecting locations only around the most semantically related reference object overly restricts the available placement choices.
The Random and Dengler~\cite{dengler2026learning} baselines show little improvement in the mean semantic score.
Their proximity scores increase modestly as the shelf becomes denser, since compact geometric placement can bring objects closer together even without explicitly optimizing semantic organization.

\subsubsection{Shelf Utilization and Placement Capacity}
Shelf density evolves similarly across the methods during the initial placement stages.
As the shelf fills, SDPP, SPS++, and Dengler achieve the highest space utilization, whereas SPS and Random terminate at substantially lower density.
Importantly, SDPP combines this dense use of the shelf with markedly better semantic organization.
Among the main methods, it achieves both the highest average shelf density and the largest average number of additional objects placed.
This demonstrates that improving semantic organization does not require sacrificing storage efficiency when spatial compactness and accessibility are considered jointly.

\subsubsection{Final Arrangement Quality}
Table~\ref{tab:placement_metrics} reports the final metrics for each run. 
In addition to the three arrangement-quality metrics, we report the number of objects placed beyond the 12 initially present and the semantic sum over all objects in the final shelf.
The semantic sum captures both the semantic quality of the arrangement and the number of successfully placed objects.
SDPP significantly outperforms all evaluated baselines in both semantic metrics.
It also achieves significantly higher shelf density than all baselines except the density-driven method of Dengler~\cite{dengler2026learning}.
Overall, SDPP attains the strongest mean performance across the placement and semantic metrics, producing arrangements that are simultaneously dense, semantically organized, and capable of accommodating more incoming objects. 

\subsubsection{Scoring-Component Ablation}
The ablation variants in Table~\ref{tab:placement_metrics} examine the effect of the AM-based accessibility-preservation term in Eq.~\ref{math:semantic_score}.
Setting $w_2=0$ removes this term while retaining the proximity-weighted semantic-density score.
This variant continues to form compact local arrangements of semantically related objects and achieves shelf density comparable to the density-driven baseline of Dengler~\cite{dengler2026learning}, while maintaining substantially better semantic organization.
Compared with this variant, the complete SDPP method places approximately 1.5 more objects per trial on average.
This suggests that local compactness alone does not fully preserve shelf regions that remain usable for subsequent placements.
Increasing the accessibility-preservation weight to $w_2=5$ yields the highest average number of additional placements and the highest shelf density, but reduces both semantic metrics.
Overall, the ablations indicate complementary roles of the two scoring components: the semantic-density term promotes compact semantic organization, whereas the accessibility-preservation term improves placement capacity by retaining usable workspace.

\subsubsection{Per-Object Planning Time}
Across the 10 trials, SDPP required $10.42 \pm 3.06\,\mathrm{s}$ per successful placement on average.
The corresponding times were $13.75 \pm 6.91\,\mathrm{s}$ for SPS, $7.12 \pm 3.34\,\mathrm{s}$ for SPS++, $14.25 \pm 2.62\,\mathrm{s}$ for Dengler, and $2.43 \pm 0.43\,\mathrm{s}$ for Random.
Although SDPP is slower than SPS++ and Random, the additional computation results in substantially better shelf arrangements.
The particularly low runtime of Random, in contrast, is accompanied by substantially lower placement capacity and arrangement quality.
Overall, these results indicate that the moderate computational overhead of SDPP provides a favorable trade-off for improved semantic organization and storage efficiency.

\subsection{Accessibility-Map Filtering Efficiency}
\label{ssec:URM_sampling}

To better understand the per-object planning times reported in Sec.~\ref{ssec:arrangement_results}, we isolate the candidate-search stage, which repeatedly invokes inverse kinematics~(IK) and motion planning until an executable placement is found.
We evaluate whether the AM-based filters can reject unlikely candidates before these expensive checks and thereby reduce the search time.

\begin{figure}[t]
    \centering
    \includegraphics[width=\linewidth]{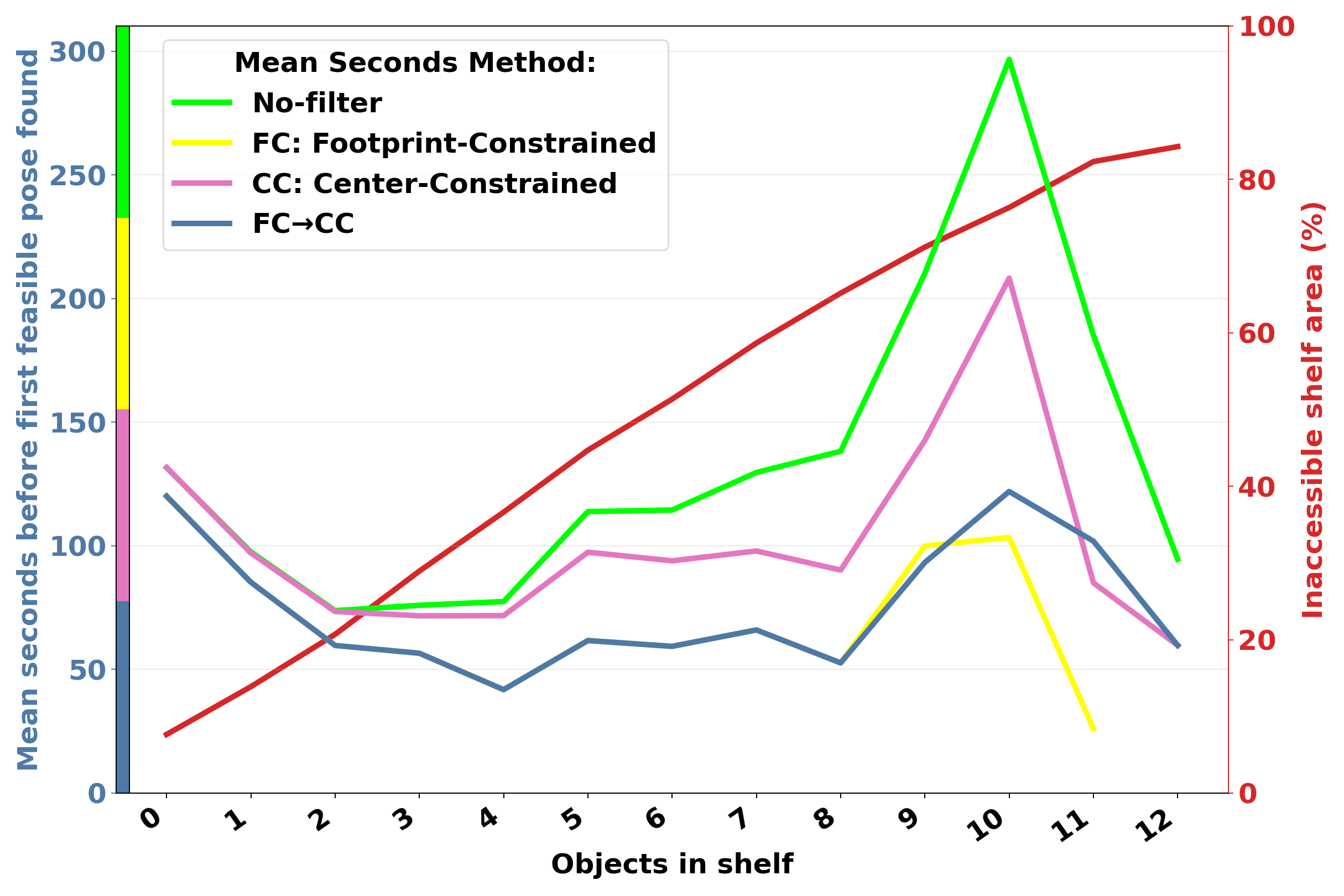}
    \vspace{-18px}
    \caption{
    Mean accumulated IK and motion-planning time required to identify the first executable placement as the shelf is progressively filled, averaged     over 20 runs.
    The no-filter baseline is shown in lime, while the AM-based strategies are shown in yellow~(FC), pink~(CC), and blue~(FC$\rightarrow$CC), where FC is followed by CC if no feasible pose is found.
    The red curve indicates the remaining accessible shelf area.
    AM-based filtering substantially reduces the candidate-search time, while FC$\rightarrow$CC retains a less restrictive fallback when FC rejects all feasible candidates.
    }
    \label{eval:AM_time_saved}
    \vspace{-12px}
\end{figure}

\subsubsection{Experimental Protocol}
Starting from an empty shelf level, we sequentially place objects using SDPP until no further valid placement can be found.
For this experiment, we additionally apply a linear depth bias toward the back of the shelf to counteract front-biased placements that can arise from semantic repulsion between dissimilar objects.
Before applying the AM-based filters, we discard candidates whose 2D projections overlap the shelf boundary or existing objects.
Boundary violations account for approximately $32.5\%$ of all sampled poses, while object-overlap rejection increases from $0\%$ to more than $80\%$ as the shelf fills.
Together, the two checks eliminate $62.35\%$ of all candidates on average and more than $85\%$ during the later placement stages.
Nevertheless, several thousand geometrically valid candidates typically remain, and we therefore evaluate a sample of m poses at each shelf state.
Candidates are evaluated either in descending order of the placement score in Eq.~\ref{math:semantic_score} or in randomized order.
Each candidate that passes the evaluated filter is first checked for IK feasibility and, if successful, subsequently evaluated using full motion planning.
We measure the accumulated IK and motion-planning time until the first executable placement is found.
Candidates rejected by a filter incur no IK or motion-planning cost in this measure.
Results are averaged over 20 independent runs.

\subsubsection{AM-Based Filter Strategies}
Geometric validity alone does not guarantee that a placement is reachable or admits a collision-free trajectory.
We therefore compare three AM-based filtering strategies with a no-filter baseline.
\textbf{Footprint-Constrained~(FC)} requires the entire candidate footprint to lie within the estimated accessible region.
This conservative criterion provides strong pruning but may reject executable placements near the boundary of the accessible region.
\textbf{Center-Constrained~(CC)} requires only the center of the candidate footprint to lie within the accessible region, while retaining the standard collision and shelf-boundary checks.
It therefore admits a larger and less conservative candidate set.
Finally, \textbf{FC$\rightarrow$CC} first applies FC, falls back to CC if no executable candidate is found, and combines the stronger pruning of FC with the greater candidate coverage of CC.

\subsubsection{Filtering Performance}
As shown in Fig.~\ref{eval:AM_time_saved}, both FC and CC substantially reduce the time required to identify the first executable placement.
However, FC occasionally over-prunes the candidate set when the shelf becomes heavily occupied.
The FC$\rightarrow$CC strategy addresses this limitation by invoking CC when FC does not yield an executable candidate, while accounting for the additional planning time incurred by the fallback.
Relative to no filtering, paired bootstrap resampling with 10,000 samples estimates mean savings of $52.8\,\mathrm{s}$ for FC$\rightarrow$CC under score-ranked ordering and $7.4\,\mathrm{s}$ under randomized ordering.
Both reductions are statistically significant according to paired $t$-tests on the run-level mean differences~($p<0.05$).
These results show that AM-based filtering reduces the expensive candidate-feasibility search under both evaluated orderings, with substantially larger savings when combined with score-based candidate ranking.

\subsubsection{Effect of Shelf Occupancy}
Figure~\ref{eval:AM_time_saved} further shows that the candidate-search time does not increase monotonically with shelf occupancy.
After the first placement, the time initially decreases because the semantic score provides stronger local guidance once objects are present, favoring placements near related objects and away from dissimilar ones.
As the shelf fills, an increasing fraction of candidates fails IK or motion planning, and the accumulated search time therefore increases, reaching a peak at approximately nine to ten objects.
At higher occupancy, more candidates are rejected by inexpensive early-stage checks, reducing the number of full motion-planning calls.
In addition, orientations close to $90^\circ$ occur more frequently and are quickly rejected when no compatible grasp is available.
This effect is particularly visible in the run reaching 12 objects.

\subsection{Qualitative Real-World Experiment}

To demonstrate the real-world applicability of our approach, we set up a UR5e robot arm next to a two-level shelf.
Object labels are obtained using SAM3~\cite{carion2025sam} with prompts adapted to the evaluated objects, as visually similar pantry items can otherwise be confused.
The accompanying video shows a qualitative run in which the robot places 24 additional objects into a shelf initially containing 9 objects.

\section{Conclusion}
\label{sec:conclusion}
In this work, we presented semantic-dense placement planning, an approach for sequential multi-object shelf storage that jointly considers semantic organization, dense space utilization, and future accessibility in confined spaces.
Experiments show that our proximity-weighted dense-semantic score with space preservation achieves significantly better semantic organization than the baselines, while also placing more objects and producing denser arrangements.
The accessibility analysis further shows that our accessibility map based filtering significantly reduces the time required to identify an executable placement, highlighting the importance of filtering unlikely candidates before motion planning.
A qualitative hardware experiment demonstrates the applicability of the proposed approach in a realistic domestic setting.

\bibliographystyle{IEEEtran}

\bibliography{bibliography}

\end{document}